\documentstyle[aclap,epsf]{article}

\long\def\comment#1{}

\newcommand{\attribute}[1]{{\sc #1}}

\newcommand{\avm}[1]{\mbox{\begin{math}
			   \setlength{\arraycolsep}{1mm}
                           \renewcommand{\arraystretch}{1}      
                           \hspace*{-0.35em} 
			   \left[ 
                           \begin{array}{@{}l@{~}l@{}} 		
                             \\[-0.16in] #1 \\[-0.16in]           
                           \end{array}
                           \right] \hspace*{-0.05em}
                           \end{math}}}

\newcommand{\tavm}[2]{\mbox{{\begin{tabular}{@{}l@{}}
                                        $\mbox{\it #1 }^{\avm{#2}}$
                                 \end{tabular}}}}

\newcommand{\avl}[2]{\mbox{\attribute{#1}} & \italic{#2}  \\ }

\newcommand{\italic}[1]{{\it #1\/}}

\newcommand{\tuple}[1]{\mbox{$\langle #1 \rangle$}}

\newcommand{\leftnel}[1]{\mbox{$\left\langle #1  \right.$}}
\newcommand{\rightnel}[1]{\mbox{$\left. #1 \right\rangle$}}

\title{\vspace{-0.5in}Applying Explanation-based Learning to
Control and Speeding-up Natural Language Generation}
\author{G\"unter Neumann \\
DFKI GmbH \\ Stuhlsatzenhausweg 3 \\ 66123 Saarbr\"ucken, Germany
 \\{\tt neumann@dfki.uni-sb.de}
}

\begin{document}
\bibliographystyle{fullname}
\maketitle
\vspace{-0.5in}
\begin{abstract}
This paper presents a method for the automatic
extraction of subgrammars to control and speeding-up natural language
generation NLG. The method is based on explanation-based learning EBL.
The main advantage for the proposed new method 
for NLG is that the complexity of the grammatical
decision making process during NLG can be vastly reduced, because the EBL
method supports the adaption of a NLG system to a particular use of a language.
\end{abstract}

\section{Introduction}
In recent years, a Machine Learning technique known as
Explanation-based Learning EBL
\cite{Mitchelletal:86,HarmelenBundy:88,Mintonetal:89} has successfully been
applied to control and speeding-up natural language parsing
\cite{Rayner:88,SamuelssonRayner:91,Neumann:94a,Samuelsson:94,SrinivasJoshi:95,RaynerCarter:96}.
The core idea of EBL is to transform the derivations (or
{\em explanations}) computed by a problem solver (e.g., a parser) to some
generalized and compact forms, which can be used very
efficiently for solving similar problems in the future. 
EBL has primarily been used for parsing to automatically specialize a
given source grammar to a specific domain.
In that case, EBL is used as a method for adapting a general grammar and/or
parser to the sub-language defined by a suitable training corpus
\cite{RaynerCarter:96}.

A specialized grammar can be seen as describing a
domain-specific set of prototypical constructions. Therefore, the EBL approach 
is also very interesting for natural language generation (NLG). 
Informally, NLG is the production of a natural language text from
computer-internal representation of information, where NLG
can be seen as a complex---potentially cascaded---decision making
process. Commonly, a NLG system is decomposed into two major
components, viz. the strategic component which decides `what
to say' and the tactical component which decides `how to say' the
result of the strategic component. 
The input of the tactical component is basically
a semantic representation computed by the strategic component. Using a
lexicon and a grammar, 
its main task is the computation of potentially all possible strings
associated with a semantic input.
Now, in the same sense as EBL is used in parsing as a means
to control the range of possible strings as well as their degree of
ambiguity, it can also be used for the tactical component
to control the range of possible semantic input and their degree of
{\em paraphrases}. 

In this paper, we present a novel method for the automatic extraction of
subgrammars for the control and speeding-up of natural language
generation. Its main advantage for NLG is that the complexity of the
(linguistically oriented) decision making process during 
natural language generation can be vastly reduced, because the EBL
method supports adaption of a NLG system to a particular language use.
The core properties of this new method are:
\begin{itemize}

\item   prototypical occuring grammatical constructions can
        automatically be extracted;
\item   generation of these constructions 
        is vastly sped up using simple but efficient mechanisms;
\item   the new method supports {\em partial} matching, 
        in the sense that new semantic input need not be
        completely covered by previously trained examples;
\item   it can easily be integrated with 
        recently developed chart-based generators as described in, e.g.,
        \cite{Neumann:94c,Kay:96,Shemtov:96}.
\end{itemize}

The method has been completely implemented and tested with a 
broad-coverage HPSG-based grammar for English (see sec. \ref{properties}
for more details).

\section{Foundations}

The main focus of this paper is tactical generation, i.e., the
mapping of structures (usually representing semantic information
eventually decorated with some functional features) to strings using a
lexicon and a grammar. Thus stated, we view tactical generation as the inverse
process of parsing. Informally, EBL can be considered as an
intelligent storage unit of example-based generalized parts of the
grammatical search space determined via training by the tactical
generator.\footnote{
In case a reversible grammar is used the parser can even be used for
processing the training corpus.}
Processing of similar new input is then reduced to simple
lookup and matching operations, which {\em circumvent} re-computation
of this already known search space.

We concentrate  on constraint-based grammar formalism following
a sign-based approach considering linguistic objects (i.e., words and
phrases) as utterance-meaning associations \cite{hpsg2}. 
Thus viewed, a grammar is a formal statement of
the relation between utterances in a natural language and
representations of their meanings in some logical or other artificial
language, where such representations are usually called {\em logical
forms} \cite{Shieber:93}. The result of the tactical generator is a
feature structure (or a set of such structures in the case of multiple
paraphrases) containing among others the input logical form, the computed
string, and a representation of the derivation.

In our current implementation we are using TDL, a typed feature-based
language and inference system for constraint-based grammars
\cite{Krieger&Schaefer:COLING-94}. TDL allows the user to define
hierarchically-ordered types consisting of type and feature
constraints. As shown later,  a systematic use of type
information leads to a very compact representation of the extracted data
and supports an elegant but efficient generalization step. 

\begin{figure*}[t]
{\small
\[
\avm{\avl{handel}{h1}
     \avl{index}{e2}
     \avl{liszt}{\leftnel{\tavm{SandyRel}{\avl{handel}{h4}
                                   \avl{inst}{x5}},
                      \tavm{GiveRel}{\avl{handel}{h1}
                                        \avl{event}{e2}
                                        \avl{act}{x5}
                                        \avl{preparg}{x6}
                                        \avl{und}{x7}},
                      \tavm{TempOver}{\avl{handel}{h1}
                                      \avl{event}{e2}}},
                      \tavm{Some}{\avl{handel}{h9}
                                  \avl{bv}{x7}
                                  \avl{restr}{h10}
                                  \avl{scope}{h11}},
                        }
    \avl{}{\rightnel{\tavm{ChairRel}{\avl{handel}{h10}
                                    \avl{inst}{x7}},
                     \tavm{To}{\avl{handel}{h12}
                                   \avl{arg}{v13}
                                   \avl{prep}{x6}},
                     \tavm{KimRel}{\avl{handel}{h14}
                                       \avl{inst}{x6}}
                    }}}
\]
}
\caption{The MRS of the string ``Sandy gives a chair to Kim''}
\label{exam-mrs}
\end{figure*}

\begin{figure*}[t]
{\small
\[
\avm{\avl{liszt}{\leftnel{\tavm{SandyRel}{\avl{handel}{h4}},
                      \tavm{GiveRel}{\avl{handel}{h1}},
                      \tavm{TempOver}{\avl{handel}{h1}},
                      \tavm{Some}{\avl{handel}{h9}},
                        }}
    \avl{}{\rightnel{\tavm{ChairRel}{\avl{handel}{h10}},
                     \tavm{To}{\avl{handel}{h12}},
                     \tavm{KimRel}{\avl{handel}{h14}}
                    }}}
\]
}
\caption{The generalized MRS of the string ``Sandy gives a chair to Kim''}
\label{exam-mrs-g}
\end{figure*}

We are adapting a ``flat'' representation of logical forms as described
in \cite{Kay:96,Copestakeetal:96}. This is a minimally structured, but
descriptively adequate means to represent semantic information,
which allows for various types of under-/overspecification, facilitates
generation and the specification of semantic transfer equivalences
used for machine translation
\cite{Copestakeetal:96,Shemtov:96}.\footnote{
But note, our approach does not depend on
a flat representation of logical forms. However, in the case of
conventional representation form, the mechanisms for
indexing the trained structures would require more complex abstract
data types (see  sec. \ref{details} for more details).}

Informally, a flat representation is obtained by the use of extra
variables which explicitly represent the relationship between the
entities of a logical form and scope information. In our current system
we are using the framework called {\em minimal recursion semantics} (MRS)
described in \cite{Copestakeetal:96}. Using their typed feature structure
notation figure \ref{exam-mrs} displays a possible MRS of the  string
``Sandy gives a chair to Kim'' (abbreviated where convenient).

The value of the feature {\sc liszt} is actually treated like a set,
i.e., the relative order of the elements is immaterial. The feature
{\sc handel} is used to represent scope information, and {\sc
index} plays much the same role as a lambda variable in conventional
representations (for more details see \cite{Copestakeetal:96}).

\section{Overview of the method}
\label{overview}

\begin{center}~
\begin{figure}[h]
        \epsffile{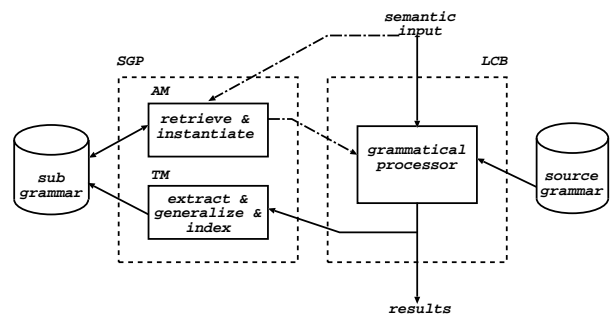}
        \caption{A blueprint of the architecture.}
\label{archi}
\end{figure}
\end{center}

The above figure  displays the overall architecture of the
EBL learning method. The right-hand part of the diagram shows the 
linguistic competence base (LCB) and
the left the EBL-based subgrammar processing component (SGP).

LCB corresponds to the tactical component of a general natural language
generation system NLG. In this paper we assume that the strategic
component of the NLG has already computed the MRS representation of the
information of an underlying computer program. 
SGP consists of a training module TM, an application module AM, and the
subgrammar, automatically determined by TM and applied by AM.

Briefly, the flow of control is as follows: During the training phase
of the system, a new logical form $mrs$ is given as input to the LCB.
After grammatical processing, the resulting feature structure $fs(mrs)$
(i.e., a feature structure that contains among others the input MRS, 
the computed string and a representation of the derivation tree)
is passed to TM. TM extracts and generalizes the derivation tree of
$fs(mrs)$, which we call the template $templ(mrs)$ of
$fs(mrs)$. $templ(mrs)$ is then stored
in a {\em decision tree}, where indices are computed from the MRS
found under the root of  $templ(mrs)$.
During the application phase, a new semantic input $mrs'$ is used for
the retrieval of the decision tree. If a candidate template can be
found and successfully instantiated, the resulting feature structure
$fs(mrs')$ constitutes the generation result of $mrs'$.

Thus described, the approach seems to facilitate only exact retrieval
and matching of a new semantic input.
However, before we describe how {\em partial
matching} is realized, we will demonstrate in more detail the exact
matching strategy using the example MRS shown in figure \ref{exam-mrs}.

\paragraph{Training phase}
The training module TM starts right after the resulting feature
structure $fs$ for the input MRS $mrs$ has been computed. In the first
phase, TM extracts and generalizes the derivation tree of $fs$, called
the template of $fs$. Each node of the template
contains the rule name used in the
corresponding derivation step and a generalization of the local MRS.
A generalized MRS is the abstraction of the {\sc liszt} value of 
a MRS where each element only contains the (lexical semantic)
type and {\sc handel} information (the {\sc handel} 
information is used for directing lexical choice (see below)).

In our example $mrs$, figure \ref{exam-mrs-g} displays
the generalized MRS $mrs_g$.
For convenience, we will use the more compact notation:

\begin{center}
\{(SandyRel h4), (GiveRel h1), \\ (TempOver h1), (Some h9),\\
(ChairRel h10), (To h12), (KimRel h14)\}
\end{center}

\begin{figure}[h]
\epsffile{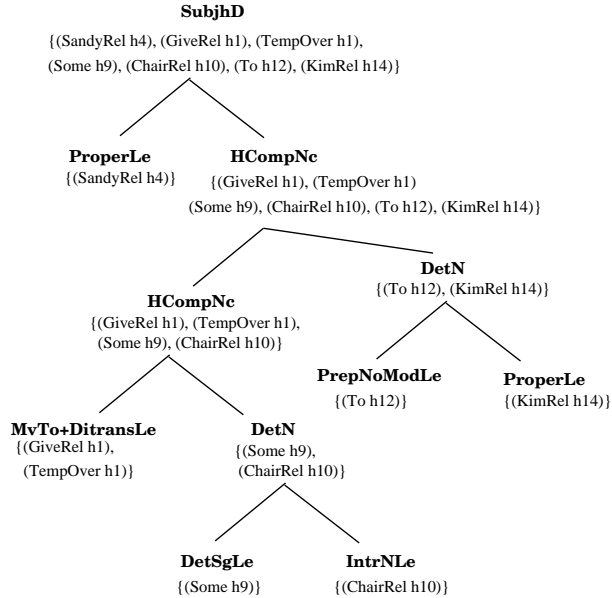}
\caption{The template  $templ(mrs)$. 
         Rule names are in bold.}
\label{dtree}
\end{figure}

\noindent Using this notation, figure \ref{dtree} (see next page) displays
the template $templ(mrs)$ obtained from $fs$.
Note that it memorizes not only the rule application structure of
a successful process
but also the way the grammar mutually relates the compositional parts
of the input MRS. 

In the next step of the training module TM, the generalized
MRS $mrs_g$  information of the
root node of $templ(mrs)$ is used for building up an index in a decision tree.
Remember that the relative order of the elements of a MRS is
immaterial. For that reason, the elements of $mrs_g$ are alphabetically
ordered, so that we can treat it as a sequence when used as a new
index in the decision tree.

The alphabetic ordering has two advantages. Firstly, we can
store different templates under a common prefix, which allows for
efficient storage and retrieval. 
Secondly, it allows for a simple efficient treatment of
MRS as sets during the retrieval phase of the application phase.

\paragraph{Application phase}
The application module AM basically performs the following steps:
\begin{enumerate}
\item   Retrieval:
        For a new MRS $mrs'$ we first construct the {\em alphabetically
        sorted} generalized MRS $mrs^{'}_{g}$. $mrs^{'}_{g}$ is then
        used as a path description for traversing the decision tree. For
        reasons we will explain soon, traversal is directed by {\em type
        subsumption}. Traversal is successful if $mrs^{'}_{g}$ has been
        completely processed and if the end node in the decision
        tree contains a template. Note that because of 
        the alphabetic ordering, the relative order of the elements of
        new input $mrs'$ is immaterial.

\item   Expansion:
        A successfully retrieved template $templ$ is expanded by
        {\em deterministically} applying the
        rules denoted by the non-terminal elements from the top
        downwards in the order specified by $templ$. 
        In some sense, expansion just re-plays the
        derivation obtained in the past. 
        This will result in a grammatically fully
        expanded feature structure, where only lexical specific
        information is still missing. But note that through structure
        sharing the terminal elements will already be constrained by
        syntactic information.\footnote{
        It is possible to perform the expansion step off-line as early
        as the training phase, in which case the application phase
        can be sped up, however at the price of more memory being taken
        up.}

\item   Lexical lookup:
        From each terminal element of the {\em unexpanded} template
        $templ$ the type and {\sc handel} information is used to select the
        corresponding element from the input MRS $mrs'$ (note that in
        general the MRS elements of the $mrs'$ are much more constrained than
        their corresponding elements in the generalized MRS
        $mrs^{'}_{g}$). The chosen input MRS element is then
        used for performing lexical lookup, where lexical elements are
        indexed by their relation name. In general this will lead to
        a set of lexical candidates.

\item   Lexical instantiation:
        In the last step of the application phase, the set of selected lexical
        elements is unified with the constraints of the terminal
        elements in the order specified by the terminal yield.
        We also call this step {\em terminal-matching}.
        In our current system terminal-matching is performed
        from left to right. Since the ordering of the terminal
        yield is given by the template, it is also possible to 
        follow other selection strategies, e.g.,
        a semantic head-driven strategy, which could lead to more
        efficient terminal-matching, because the head element is
        supposed to provide selectional restriction information for its
        dependents.
       
\end{enumerate}

A template together with its corresponding index describes all sentences of
the language that share the same derivation and whose MRS are
consistent with that of the index. 
Furthermore, the index and the MRS of a template together
define a normalization for the permutation of the elements of a new input MRS.
The proposed EBL method guarantees {\em soundness} because retaining and
applying the original derivation in a template enforces the full
constraints of the original grammar.

\paragraph{Achieving more generality}
So far, the application phase will only be able to re-use templates for
a semantic input which has the same semantic type information. However, it
is possible to achieve more generality, if we apply a further
abstraction step on a generalized MRS. This is simply achieved by
selecting a {\em supertype} of a MRS element instead of the given
specialized type. 

The type abstraction step is based on the standard
assumption that the word-specific lexical semantic types can be
grouped into classes representing morpho-syntactic paradigms. 
These classes define the upper bounds for the abstraction
process. In our current system, these upper bounds are directly 
used as the supertypes to be considered during the type abstraction step.
More precisely, for each element $x$ of a generalized MRS $mrs_g$
it is checked whether its type $T_x$ is subsumed by an upper bound  $T_s$
(we assume disjoint sets). Only if this is the case, 
$T_s$ replaces $T_x$ in $mrs_g$.\footnote{
Of course, if a very fine-grained lexical semantic type hierarchy is
defined then a more careful selection would be possible to obtained
different degrees of type abstraction and to achieve a more
domain-sensitive determination of the subgrammars. However, more
complex type abstraction strategies are then needed which would be able to
find appropriate supertypes automatically.
}
Applying this type abstraction strategy
on the MRS of figure \ref{exam-mrs}, we obtain:

\begin{center}
\{(Named h4), (ActUndPrep h1), \\ (TempOver h1), (Some h9),\\
(RegNom h10), (To h12), (Named h14)\}
\end{center}

\begin{figure}[h]
\epsffile{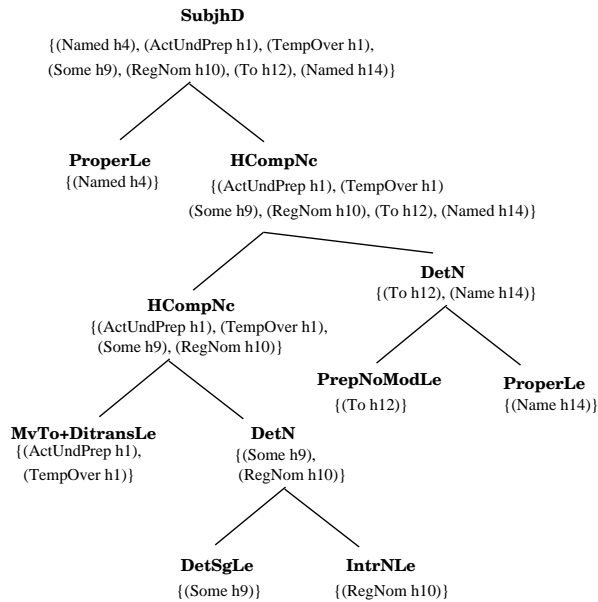}
\caption{The more generalized derivation tree $dt_g$ of $dt$.}
\label{dtree2}
\end{figure}

\noindent where e.g., {\sc Named} is the common supertype of {\sc SandyRel} and
{\sc KimRel}, and {\sc ActUndPrep} is the supertype of {\sc GiveRel}. Figure
\ref{dtree2} shows the template $templ_g$ obtained from $fs$
using the more general MRS
information. Note, that the MRS of the root node is used for building
up an index in the decision tree.

Now, if retrieval of the decision tree is directed by type subsumption,
the same template can be retrieved and potentially instantiated
for a wider range  of new MRS input, namely for those which are 
{\em type compatible} wrt. subsumption relation. Thus, the template
$templ_g$ can now be used  to generate, e.g., the string ``Kim gives a
table to Peter'', as well as the string ``Noam donates a book to
Peter''.

However, it will not be able to generate a sentence like ``A man
gives a book to Kim'', since the retrieval phase will already fail.
In the next section, we will 
show how to overcome even this kind of restriction.

\section{Partial Matching}
\label{details}

The core idea behind partial matching is that in case an exact
match of an input MRS fails we  want at least as many subparts as possible 
to be instantiated. Since the instantiated template of a MRS
subpart corresponds to a phrasal sign, we also call it a {\em phrasal
template}. For example, assuming that the training phase has
only to be performed for the example in figure \ref{exam-mrs}, 
then for the MRS of ``A man gives a book to Kim'', a partial match
would generate the strings ``a man'' and ``gives a book to
Kim''.\footnote{
If we would allow for an exhaustive partial match (see below) then
the strings ``a book'' and ``Kim'' would additionally be generated.}
The instantiated phrasal templates are then combined by the tactical
component to produce larger units (if possible, see below).

\paragraph{Extended training phase}
The training module is adapted as follows:
Starting from a template $templ$ obtained for the training
example in the manner described above, 
we extract recursively all possible subtrees $templ_s$ also
called {\em phrasal templates}. Next,
each phrasal template is inserted in the decision tree in the way
described above. 

It is possible to direct the
subtree extraction process with the application of {\em filters},
which are applied  to the whole remaining subtree in each recursive step. 
By using these filters it is possible to restrict the range of
structural properties of candidate phrasal templates (e.g., extract
only saturated NPs, or subtrees having at least two daughters, or subtrees
which have no immediate recursive structures). These filters serve the
same means as the ``chunking criteria'' described in \cite{RaynerCarter:96}.

During the training phase it is recognized for each phrasal template $templ_s$
whether the decision tree already contains a path 
pointing to a previously extracted and already stored phrasal template
 $templ^{'}_{s}$, such that $templ_s = templ^{'}_{s}$. 
In that case, $templ_s$ is not inserted and the recursion stops
at that branch.

\paragraph{Extended application phase}

For the application module, only the retrieval operation of the
decision tree need be adapted. 

Remember that the input of the retrieval operation is the sorted
generalized MRS $mrs_g$ of the input MRS $mrs$. Therefore, $mrs_g$
can be handled like a sequence. The task of the
retrieval operation in the case of a partial match is now to
potentially find all subsequences of $mrs_g$ which lead to a template.

In case of exact matching strategy, the decision tree must be visited
only once for a new input. In the case of partial matching, however, the
decision tree describes only possible {\em prefixes} for a new
input. Hence, we have to recursively repeat retrieval of the  decision tree
as long as the remaining suffix is not empty. 
In other words, the decision tree is now a
finite representation of an infinite structure, because implicitly,
each endpoint of an index bears a pointer to the root of the
decision tree.

Assuming that the following template/index pairs have been inserted into
the decision tree: \tuple{ab,t_1}, \tuple{abcd,t_2}, \tuple{bcd,t_3}.
Then retrieval using the path $abcd$ will return all three templates,
retrieval using $aabbcd$ will return template $t_1$ and $t_3$, and
$abc$ will only return $t_1$.\footnote{
It is possible to parameterize our system to perform an {\em
exhaustive} or a {\em non-exhaustive} strategy. In the non-exhaustive
mode, the longest matching prefixes are preferred.}

\paragraph{Interleaving with normal processing}
Our EBL method can easily be integrated with normal processing,
because each instantiated template can  be used directly as an 
already found sub-solution. In case of an agenda-driven chart generator
of the kind described in \cite{Neumann:94a,Kay:96}, an instantiated
template can be directly added as a {\em passive edge} to the
generator's agenda. If passive edges with a wider span are given higher
priority than those with a smaller span, the tactical generator would try to
combine the largest derivations before smaller ones, i.e., it would
prefer those structures determined by EBL.

\section{Implementation}
\label{properties}
The EBL method just described has been fully implemented and tested
with a broad coverage HPSG-based English grammar including more than
2000 fully specified lexical entries.\footnote{This grammar has been 
developed at CSLI, Stanford, and kindly be provided to the author.}
The TDL grammar formalism is very powerful, supporting distributed
disjunction, full negation, as well as full boolean type logic.

In our current system, an efficient chart-based bidirectional
parser is used for performing the training phase. 
During training, the user can interactively select which of the
parser's readings should be considered by the EBL module. In this way the
user can control which sort of structural ambiguities should be avoided because
they are known to cause misunderstandings.
For interleaving the EBL application phase with normal
processing a first prototype of a chart generator has been implemented
using the same grammar as used for parsing. 

First tests has been carried out using a small test set of 179
sentences. Currently, a parser is used for processing the test set
during training. Generation of the extracted templates is performed
solely by the EBL application phase (i.e., we did not considered
integration of EBL and chart generation). 
The application phase is very efficient. 
The average processing time for indexing and instantiation of
a sentence level template (determined through parsing)
of an input MRS is approximately one
second.\footnote{
EBL-based generation of all possible templates of an input MRS 
is less than 2 seconds. The tests have been performed using a Sun UltraSparc.
}
Compared to parsing the corresponding string the factor of speed up is
between 10 to 20. A closer look to the four basic EBL-generation steps:
indexing, instantiation, lexical lookup, and terminal matching showed
that the latter is the most expensive one (up to 70\% of computing
time). The main reasons are that 1.) lexical lookup often returns
several lexical readings for an MRS element (which introduces lexical
non-determinism) and 2.) the lexical
elements introduce most of the disjunctive constraints which makes
unification very complex. Currently, terminal matching is performed left
to right. However, we hope to increase the efficiency of this step by
using head-oriented strategies, since this might help to re-solve
disjunctive constraints as early as possible. 

\section{Discussion}
\label{discussion}
The only other approach I am aware of which also considers EBL for NLG
is \cite{Samuelsson:95a,Samuelsson:95b}. However, he focuses on the
compilation of a logic grammar using LR-compiling techniques, where
EBL-related methods are used to optimize the compiled LR tables, in
order to avoid spurious non-determinisms during normal generation. 
He considers neither the extraction of a specialized grammar for supporting
controlled language generation, nor strong integration with
the normal generator. 

However, these properties are very important for achieving
high applicability. Automatic grammar extraction is worthwhile because
it can be used to support the definition of a {\em controlled} domain-specific
language use on the basis of training with a general source grammar.
Furthermore, in case exact matching is requested
only the application module is needed for processing the
subgrammar.
In case of normal processing, our EBL method serves as a speed-up
mechanism for those structures which have ``actually been used or
uttered''. However, completeness is preserved.

We view generation systems which are based on ``canned text'' and 
linguistically-based systems simply as two endpoints of a
contiguous scale of possible system architectures
(see also \cite{Daleetal:94}). 
Thus viewed, our approach is directed towards the
automatic creation of application-specific generation systems.
        
\section{Conclusion and Future Directions}
\label{future-work}
We have presented a method of automatic extraction of subgrammars for
controlling and speeding up natural language generation (NLG). 
The method is based on explanation-based learning (EBL), which has already
been successfully applied for parsing. We showed how the method can
be used to train a system to a specific use of grammatical and lexical
usage.

We already have implemented a similar EBL method for parsing, which
supports on-line learning as well as statistical-based management of
extracted data.
In the future we plan to combine EBL-based generation and parsing
to one {\em uniform} EBL approach usable for high-level performance
strategies which are based on a strict interleaving of parsing and
generation (cf. \cite{NeumannVanNoord:94c,Neumann:94a}).

\section{Acknowledgement}
The research underlying this paper was
supported by a research grant from the German
Bundesministerium f\"ur Bildung, Wissenschaft, Forschung und
Technologie (BMB+F) to the DFKI project {\sc paradime} FKZ~ITW~9704.

I would like to thank the HPSG people from CSLI, Stanford
for their kind support and for providing the HPSG-based English
grammar. In particular I want to thank Dan Flickinger and Ivan Sag.
Many thanks also to Walter Kasper for fruitful discussions.


\end{document}